\documentclass{article}

\usepackage{PRIMEarxiv}

\usepackage[utf8]{inputenc} 
\usepackage[T1]{fontenc}    
\usepackage{hyperref}       
\usepackage{url}            
\usepackage{booktabs}       
\usepackage{amsfonts}       
\usepackage{nicefrac}       
\usepackage{microtype}      
\usepackage{lipsum}
\usepackage{fancyhdr}       
\usepackage{graphicx}       
\graphicspath{{media/}}     
\usepackage{listings}
\usepackage[T1]{fontenc}
\usepackage{graphicx}
\usepackage{adjustbox}
\usepackage{xcolor} 
\title{Feedback Indicators: The Alignment between Llama and a Teacher in Language Learning
}

\author{
  Sylvio Rüdian \\
  Humboldt-Universität zu Berlin \\
  Department of Computer Science \\
  Berlin, Germany\\
  \texttt{\ 0000-0003-3943-4802} 
  \And 
  Yassin Elsir \\
  Humboldt-Universität zu Berlin \\
  Department of Computer Science \\
  Berlin, Germany\\
  \texttt{\ 0009-0000-4747-7920} \\
  \And 
  Marvin Kretschmer \\
  Humboldt-Universität zu Berlin \\
  Department of Computer Science \\
  Berlin, Germany\\
  \texttt{\ 0009-0008-5560-8255} \\
  \AND 
  Sabine Cayrou \\
  Humboldt-Universität zu Berlin \\
  Language Centre \\
  Berlin, Germany\\
  \texttt{\ 0009-0004-0288-0715} \\
  \And 
  Niels Pinkwart \\
  German Research Center \\
for Artificial Intelligence (DFKI) \\
  Berlin, Germany\\
  \texttt{\ 0000-0001-7076-9737} \\
}

\begin{document}
\maketitle

\begin{abstract}
Automated feedback generation has the potential to enhance students' learning progress by providing timely and targeted feedback. Moreover, it can assist teachers in optimizing their time, allowing them to focus on more strategic and personalized aspects of teaching. To generate high-quality, information-rich formative feedback, it is essential first to extract relevant indicators, as these serve as the foundation upon which the feedback is constructed. Teachers often employ feedback criteria grids composed of various indicators that they evaluate systematically. This study examines the initial phase of extracting such indicators from students' submissions of a language learning course using the large language model Llama 3.1. Accordingly, the alignment between indicators generated by the LLM and human ratings across various feedback criteria is investigated. The findings demonstrate statistically significant strong correlations, even in cases involving unanticipated combinations of indicators and criteria. The methodology employed in this paper offers a promising foundation for extracting indicators from students' submissions using LLMs. Such indicators can potentially be utilized to auto-generate explainable and transparent formative feedback in future research.
\end{abstract}

\keywords{Large Language Models \and Prompt Engineering \and Feedback Indicators \and Language Learning.}

\section{Introduction}
Feedback plays an essential role in education due to the substantial potential impact on learners' development and learning success \cite{hattie2009visible}. Implemented appropriately, it can increase the students' effort and engagement and help them develop better strategies to improve their understanding \cite{hattie2007power}. For this purpose, according to Hattie and Timperley \cite{hattie2007power}, it should provide information about the learning goals, the current progress of the student, and possible next learning steps. Here, high-information feedback, i.e., information about the correctness of the student response, the learning process itself, and the student's self-regulation, has been shown to be incredibly effective at improving learning results \cite{wisniewski2020power}. Due to its validity and reliability, an established method for providing feedback is a criterion-referenced approach, in which learners' performance is assessed according to predefined criteria \cite{burton2005design}. Another effective feedback modality for enhancing the learning process is formative feedback \cite{irons2021enhancing}. There, students receive regular updates concerning their performance and are given guidance on how to achieve the learning objectives best. McCallum and Millner \cite{mccallum2021effectiveness} indicated that students found this type of feedback motivating and helpful for monitoring their progress and increasing their perceived level of understanding. Furthermore, the practice also benefits teachers, who can use the diagnostic information to improve their teaching and provide targeted student support.

These benefits of effective feedback are particularly useful in the context of language learning. Language skills remain a necessary inter-professional key qualification for lifelong learning in an increasingly globalized environment \cite{eu2019keycompetences}. Language teaching currently relies on the CEFR Companion Volume \cite{CEFRCompanionvolume}, which emphasizes an action-oriented approach \cite{CEFRaction-orientedapproach}, including mediation and online interaction as well as plurilingual and -cultural aspects. Therefore, it aims to enable each student to act as a social agent in various contexts in a linguistically and culturally confident and appropriate way. Since developing language skills involves complex and abstract mental processes in learners \cite[p. 278]{vanpatten2020theories}, it is essential to consider the entire learning process when providing feedback. Teachers should employ a diverse set of feedback strategies, adopting the role of a learning coach who is attentive to the individual needs of each student \cite{nassaji2016anniversary}. Continuous dialogue between teacher and student plays a supportive role in the learning process as it stimulates cognitive engagement and encourages students to reflect on their progress and control their learning paths \cite{kleppin2019korrektivesfeedback}. However, providing personalized, continuous, and high-information feedback is time-consuming for teachers \cite{henderson2019challenges}. Due to contextual constraints on teaching resources, providing detailed regular feedback can be challenging, especially in higher education, where student cohorts are often quite large \cite{henderson2019challenges}. The circumstance that teachers of large classes regularly lack detailed knowledge of students further complicates the provision of adequate feedback, which can be challenging.

Automated feedback systems, technologies within digital learning environments that automatically provide students with individual feedback, have the potential to address these challenges by delivering detailed feedback en mass and reducing the workload \cite{deeva2021review}. At the same time, they can offer further educational advantages by reducing biases and increasing grading consistency \cite{hahn2021systematic}. Here, natural language processing (NLP) and other machine learning methods can be used to extract valuable information from historical data and improve these systems. While most automated feedback systems address STEM education, language learning is also a common area of application \cite{deeva2021review}. In this field, the release of capacities by these systems may allow a shift from language teaching to a student-centered coaching model with personalized, optimized, technology-enhanced learning. Here, integrating artificial intelligence (AI) into assessment and feedback provided by language teachers is still an emerging field \cite{Liu2023AIlanguageteaching}. Vogt and Flindt \cite{vogt&flindt2023AItoolslanguageeducation} highlight the growing trend of using AI applications for personalized learning in Germany. However, research on the impact of AI-powered tools in foreign language education remains limited \cite{vogt&flindt2023AItoolslanguageeducation}.

A notable challenge in implementing AI-supported feedback systems in certain educational settings is the scarcity of data. Besides large lectures, many courses in higher education contain smaller participant numbers, which nevertheless entail a high correction effort for the lecturers due to regular submissions. The reduced number of participants limits the available data for these courses, which is critical for developing AI applications. Addressing this challenge necessitates specialized procedures tailored for this purpose \cite{nguyen2024learning}.

An upcoming technology that offers a lot of potential for data-driven automatic feedback systems is a Large Language Model (LLM). Such models have the capability to generate text responses based on textual instructions \cite{ouyang2022training} and are suitable for a variety of language-related tasks \cite{kasneci2023chatgpt}, like text summarization \cite{jin2024comprehensive} or rewriting \cite{shu2024rewritelm}. Despite recognized limitations in abstract reasoning \cite{wu2023reasoning} or the potential of hallucinations \cite{zhang2023sirens}, LLMs demonstrate promising results in open domains. For this reason, attempts are currently being made to utilize this potential in the field of education and to integrate LLMs in various applications \cite{wang2024largelanguagemodelseducation}. In this paper, the domain of language teaching is chosen, employing LLMs to enhance the feedback generation process.

\section{Related Work}
AI, as a disruptive technology, is changing the way we learn, above all in the field of foreign language learning. AI writing assistants can support students in reflecting on their writing process and help them to identify and correct grammatical errors \cite{Hartmann2021disruptivetechnologiesGFLteaching}. Godwin-Jones \cite{Godwinjones2022partneringwithAI} distinguishes between two types of AI writing tools: those designed for academic purposes, known as Automated Writing Evaluation (AWE), and those used in educational or professional settings, referred to as Automated Written Corrective Feedback (AWCF). Unlike AWE tools, AWCF tools focus on lower-level writing issues, such as grammatical and lexical errors \cite[p. 7-8]{Godwinjones2022partneringwithAI}. The accessibility and, in some cases, widespread use of these tools, along with the immediate feedback they provide through error classification, are seen as advantages \cite{O’NeillRussell2019grammarchecker,schmidtstrasser2022AIintelligentpractice}. Such feedback can support students to become more aware of their writing output and improve their lexical variation and overall writing skills \cite{dizon2021examininggrammarly}.

Kahneman \cite{Kahneman2011Thinkingfastandslow} reports that developing intuitive expertise depends on the quality and speed of feedback and sufficient opportunity to practice. 
Implementing such tools in the English as a Foreign Language classroom can reduce the teachers' workload in providing feedback, allowing them to focus on other instructional tasks, such as supporting students in utilizing the feedback. However, employed tools cover only a fraction of feedback criteria. Subsequently, providing feedback manually is still required.
Strasser \cite{Strasser2020feedbackKItools} notes that AI-supported feedback has already become very user-friendly. The advantage of automated test evaluation is the aspect of immediacy. Such tools can provide feedback in seconds. Thanks to this quick response time, learners can take immediate action to correct their linguistic insufficiencies and improve their skills \cite{teske2017duolingo}. Exemplary, the language learning platform \textit{Parkur} \cite{parkurofaj} has demonstrated an example of effectively combining auto-corrective learning settings with individual formative feedback through human online tutoring \cite{Segura2022parkurofaj}. Students receive immediate textual corrective feedback automatically after having completed the task. There, AI-supported feedback could beneficially support students and teachers. 

AI methods have expanded the possibilities for feedback generation within automated feedback systems. In their review, Deeva et al. \cite{deeva2021review} identified three approaches to create feedback generation models: expert-driven, data-driven, and mixed approaches. Expert-driven approaches rely only on expert knowledge, typically comparing predefined answers 
with sample solutions or similar, and evaluating them using simple rule-based algorithms. In contrast, data-driven approaches derive their evaluation models solely from student data using AI techniques. Last, mixed approaches represent a combination of both expert-driven and data-driven elements. Therefore, both types of systems, which employ data-driven and mixed methodologies, generate feedback supported by AI.

Feedback systems can use automated extraction procedures to parse indicators from student submissions, subsequently used to train models, and produce the predictions for texts that form the basis of feedback \cite{rudian2023pre}. The advent of modern natural language processing (NLP) has enabled the extraction of complex features from texts \cite{nelson2022bridginghumanintelligence}.NLP techniques rely on computational analysis of semantics and syntax \cite{nelson2022bridginghumanintelligence} to autonomously extract information from text. Traditionally, NLP-based information extraction (IE) has been accomplished by rules-based and statistical methods \cite{nelson2022bridginghumanintelligence}, which parse information from texts based on a set of pre-defined extraction rules or a probability distribution, enabling the identification of representative patterns within texts. However, the heuristic methods are limited by the necessity to develop accurate rules for each novel feature 
\cite{kiran2019limitationsofinformationextractionmethods}. In contrast, machine learning methods,  utilizing probabilistic language models, can be setup to reliably extract novel features without prior training or rule development \cite{yang2022asurveyofinformationextraction}. 

DL-IE has reached new heights with the advent of LLMs. They demonstrate strong results on benchmarks measuring natural language understanding across various professional and academic domains \cite{Hendrycks2020MeasuringMM}. In particular, the strong performance of models such as Llama 3 \cite{Dubey2024TheL3} suggests a high potential for autonomous extraction of complex information. The suitability of LLMs for information extraction has been demonstrated in a number of recent studies, for example, in extracting information from scientific literature \cite{thomas2024large,polak2024extractingaccuratematerialsdata}.
Thomas et al. \cite{thomas2024large} investigated the use of an LLM for search engine content relevance labeling. The LLM has been induced to produce content labels via a handcrafted prompt consisting of a detailed instruction describing the task and the required content information. The researchers were able to iteratively improve the accuracy in terms of relevancy, resulting in a high degree of alignment between LLM and human graders. Polak et al. \cite{polak2024extractingaccuratematerialsdata} introduce an approach that engineers an entire conversation with an LLM, whereby data is extracted, classified, and evaluated. The prompt technique utilizes queries optimized for each possible scenario to force the LLM interaction into predictable patterns, allowing the researchers to produce deterministic outcomes. However, such an approach requires careful validation of the potential for variation in prompt-response pairings to avoid unforeseen negative consequences.

As demonstrated by Thomas et al. \cite{thomas2024large} and Polak et al. \cite{polak2024extractingaccuratematerialsdata}, the state-of-the-art has highlighted the challenges facing automated feedback generation. The autonomous generation of feedback necessitates techniques that can extract basic or complex features from text submissions without a human-in-the-loop and without compromising feedback integrity. To that end, major research gaps include the appropriate query design and model choice for the class of IE task and techniques for effectively evaluating generated results. A thorough classification of generation techniques and methods would be of benefit to researchers seeking to automate feedback processes. Similarly, techniques for validating LLM extraction processes would be beneficial for integrating automated feedback into the resources available to support student development. 

Feedback generation relying solely on LLMs faces some limitations. Namely, the manner of LLM responses may be generic, hallucinated, or irrelevant. For example, Estévez-Ayres et al. \cite{estevez2024evaluation} investigated whether LLMs are reliable when providing feedback on concurrency programming errors. The feedback generated by both GPT-3.5 and PaLM 2 provided useful responses at a rate of 50\%, utilizing instructor feedback as a baseline. An increasingly popular approach in the context of automated assessment systems, which could address these limitations, is the use of LLMs to evaluate open responses based on predefined criteria. Mizumoto and Eguchi \cite{mizumoto2023exploring} used a GPT-3.5 model to score essays by non-native English learners from the TOEFL11 corpus \cite{blanchard2014ets} 
based on linguistic criteria within the scale of 0 to 9. The study found that the results of the LLM closely aligned with the gold standard, which is a promising indication of the model’s accuracy. Similar approaches to generate scores based on predefined criteria were done in the context of mathematics tasks. Lee et al. \cite{lee2024can} investigated the ability of GPT-4 to assess open-ended mathematics questions for sixth-grade students and found a statistically significant correlation with the ratings of human evaluators for most tasks. Meanwhile, Gandolfi \cite{gandolfi2024gpt} obtained a promising result when using GPT-3.4 and GPT-4 to evaluate calculus exercises. However, the level of a human evaluator could not be achieved. In addition, problems such as simple arithmetic errors and hallucinations were still observed. 
To increase transparency and generate explainable feedback for students, LLMs' capabilities can be leveraged to extract relevant information from texts for which traditional NLP approaches have already reached their limits. These indicators can be used to train white-box models such as regression trees, which are able to accurately and transparently predict the scores of texts based on the extracted indicators \cite{rudian2023pre}. This predictive capacity can be utilized to supplement student learning by providing automated feedback. In this paper, we examine whether linguistic and contextual indicators extracted from student submissions employing LLMs correlate with competency-based teacher ratings. If moderate to high correlations can be identified, such indicators can be a fruitful source to assist teachers when rating student submissions. To the best of the authors' knowledge, researchers have still missed the intermediate step of automatically extracting relevant indicators from texts that align with teachers' feedback. This paper aims to bridge that gap.

\section{Methodology}
This section describes the online course used for the study. The course generates data through submission and feedback, which is subsequently utilized in the study to develop and assess automated generation techniques.
\subsection{The Online Course}
An online course was developed to teach French as a foreign language (level B1), based on the principles of Biggs' Constructive Alignment \cite{biggs1996constructivealignment}. The structure and course content were aligned with the learning objectives and the examination tasks, based on the internally defined evaluation criteria of UNIcert® I at level B1 of the CEFR \cite{council2001common}. The learning objective set out for students was to reach the B1 level by the end of the course.
In the online course, students were asked to complete tasks to improve their writing skills voluntarily. The tasks consisted of short narratives of varying lengths and utilized the taught grammatical structures and learned vocabulary. The voluntary tasks were submitted to a forum moderated by the teacher, where they received regular feedback to improve submission quality. Providing textual formative feedback regularly helped to track the student's learning progress and identify frequent mistakes to adapt the course. In the course, learning content has been repeated if necessary. Over the semester, learners completed micro-tasks for each lesson (lexis, grammatical structures, reading and listening tasks), to prepare for a collaborative final macro-task. According to Cayrou \cite{Cayrou2021DicoolB1Frenchcourse}, a language course in which learners had regularly written short narratives and received formative feedback leads to more fluent interactions. Our experiment was conducted in the form of micro-tasks. Students were asked to describe a movie they could remember well by using adjectives and relative pronouns to describe their impressions and feelings. Thereafter, students were to discuss a movie recommendation. Teachers provided feedback via a feedback grid implemented with the Moodle Plug-in \textit{Rubric}, inspired by a criteria grid that Jaeger has adapted  \cite{Jaeger2017interculturel} (guided by Tagliante's \cite{Tagliante2016CECR} work for language skills evaluation). The feedback grid consisted of the following criteria: (1) the consideration of the task, (2) sociolinguistic appropriateness, (3) information and description skills, (4) general mediation skills, (5) coherence and cohesion, (6) range of vocabulary, (7) mastery of vocabulary, (8) mastery of spelling, (9) grammatical correctness, and (10) morphosyntax.
\subsection{The Study}
In the study, we aim to identify LLM-generated values for feedback indicators of texts submitted by language learners that align with teacher ratings. First, based on the provided criteria descriptions \cite{Jaeger2017interculturel}, a set of indicators was identified that could represent a criterion in the feedback grid. For example, the description of sociolinguistic appropriateness: ``Can realize and respond to a wide range of language functions using the most common means of expression and a neutral register. Politeness conventions are mastered as well as customs, attitudes, values, and beliefs in the society concerned and their signals.'' can be utilized to extract an indicator of the relevant indicator score from a text.
\cite{Jaeger2017interculturel} leads to a set of textual indicators that is extracted from the description to ask whether the learners used the most common means of expression, neutral register, politeness conventions (form of address, formality, polite expressions), and values and beliefs. The existence of each attribute must be validated for all learner submissions. To this end, we have designed a set of prompts to employ an LLM to extract the indicator score for each attribute from a given text. Llama3.1 is utilized for all experiments \cite{touvron2023llama}. Prompts fell into two classes: binary and list. Binary prompts were created, which returned Yes/No responses (e.g., whether a formal address was used). Similarly, list prompts return a response in a list format  (e.g., all formal expressions are extracted, where the number of such expressions is further processed). For example, the prompt ``Does this text use a form of address?'' returns a binary result, indicating whether a form of address is included. For such results, ``Respond exclusively with YES for yes and NO for no.'' has been added to restrict the output formatting. When asking for the number of certain elements, the prompt ``Do not list any errors. Return the list in JSON format {key:data}'' is added, and from that, the number of returned elements is counted to suit as an indicator. 
With them, numbers for each textual indicator can automatically be received. However, to be helpful in an educational setting, whether such LLM-generated indicator values are realistic and correct must be evaluated. Teachers can directly assess the LLM-generated indicator values for each text \cite{rudian2023auto}. This type of evaluation requires human resources for manual assessment. Alternatively, LLM-generated values of indicators can be compared to criteria that teachers have already rated in historical learner submissions \cite{Rüdian2024e}. This indirect evaluation has the advantage that existing textual learner submissions, along with ratings from a feedback criteria grid, can be used directly without the need for additional human resources. This paper applies the indirect human alignment between LLM-generated feedback indicators and teacher ratings. The resulting indicator set can then be utilized to train explainable supervised regression models, which can subsequently be employed to predict feedback criteria ratings.

\section{Results}
22
students enrolled in the course during the winter semester 2023-24. The estimated duration of the course was 60 hours, distributed over a period of 15 weeks. The course took place twice a week and addressed a heterogeneous group of students of all faculties. Of the 
22
students, 
11
22 completed the analyzed task and gave their consent to process their submissions, including feedback. 

Next, correlation analysis was conducted. We identified 44 of 63 extracted indicators (70\%) resulting an acceptable Pearson correlation coefficient with at least one criterion ($|PCC|\geq .5$). Specifically, this was observed for five of the 11 binary indicators (45\%) and 39 of the 52 list indicators (75\%). Of these 44 indicators, 15 demonstrated high correlation to at least one evaluation criterion with $|PCC|\geq.70$, that have been statistically significant (.01 level, two-tailed), as shown in Table \ref{fig:fig1}. This was the case for three binary (18\%) and 12 of the list indicators (23\%). 34 relevant correlations between indicators and criteria have been identified, including 10 with a high statistically significant correlation ($|PCC|\geq.8$).

It can be observed that individual indicators usually showed such high correlations for multiple criteria, with a median of two criteria per indicator. 
Conversely, each criterion demonstrated a highly significant correlation with at least one indicator, with a median of 3 indicators per criterion. 
In the experiment, most of these indicators demonstrated a highly significant correlation concerning the criterion "information and description skills" (10 indicators). Additionally, some features highly correlated with the criteria ``sociolinguistic appropriateness'' and ``range of vocabulary'' (4 indicators). The criteria ``coherence and cohesion'' and ``consideration of the task'' each exhibited only one indicator with a high significant correlation, with the latter being the only criterion that had no significant correlation with any indicator ($|PCC|\geq.8$), which is not surprising, as no prompt has been prepared for that criterion due to imbalanced data.


\begin{table}[!ht]
\centering
\setlength{\tabcolsep}{0pt} 
\renewcommand{\arraystretch}{0} 
\begin{tabular}{@{}p{\textwidth}@{}}
\adjustbox{valign=c}{
    \includegraphics[clip, trim=3cm 2.5cm 4cm 2cm,width=.9\textheight, angle=270,origin=c]{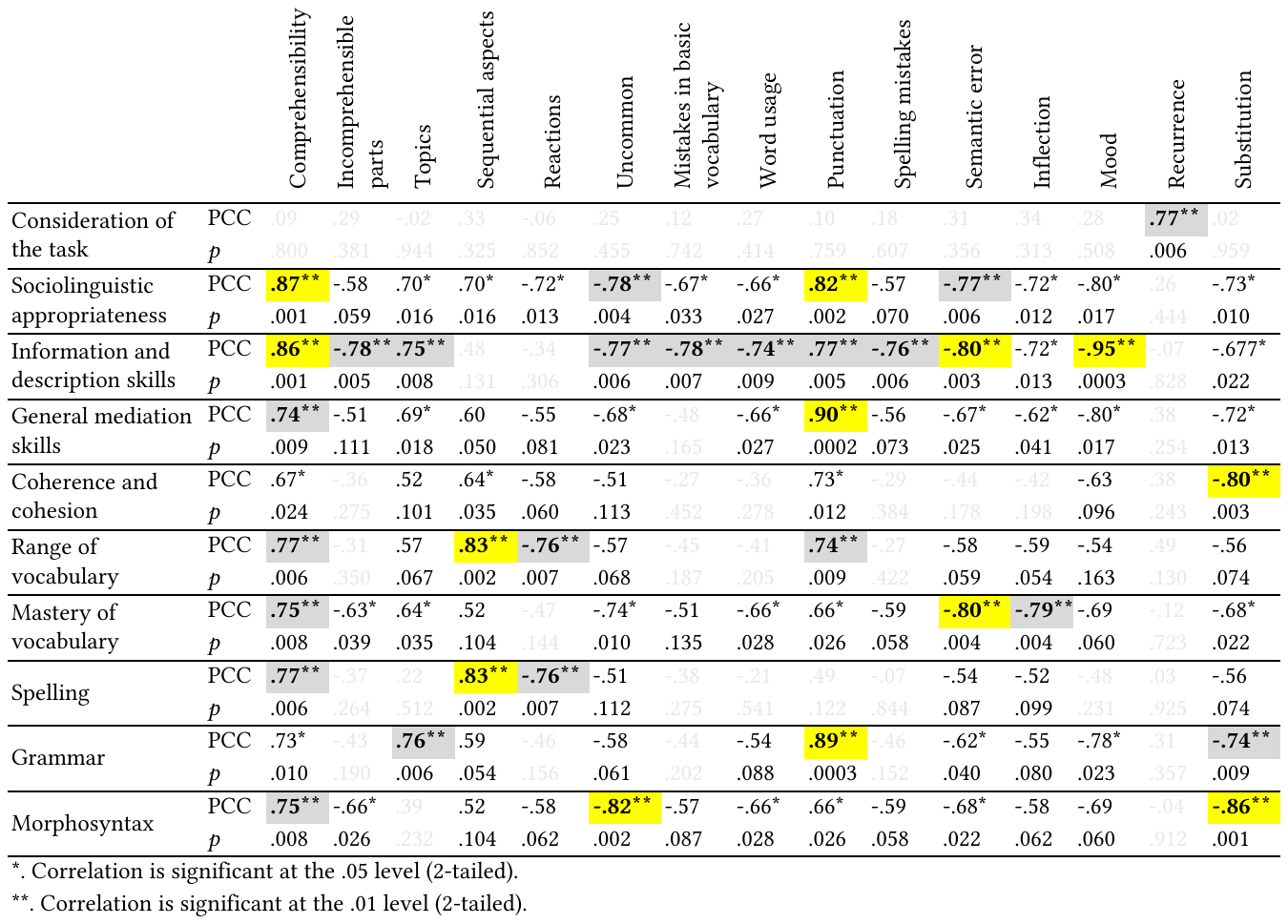}
}
\end{tabular}
\caption{Correlation Analysis between competencies and indicators. $PCC$: Pearson's Correlation Coefficient; $p$: significance (2-tailed). Values are in bold if $p\leq.01$ (**), with visual highlights applied when 
    \colorbox{gray!30}{$|PCC|\geq.70$ (gray)} 
    or 
    \colorbox{yellow!100}{$|PCC|\geq.8$ (yellow)}. Light grey font indicates values not of interest $(|PCC|<.5)$.
    }

\label{fig:fig1}
\end{table}

\section{Discussion}
The results revealed moderate to high correlations between feedback indicators generated by the LLM and teacher ratings for several feedback criteria. This promising finding allows us to utilize a pre-defined set of indicators that an LLM can process to obtain numerical values, which can then be further processed to predict teacher ratings on several feedback criteria using a supervised approach.

Despite observing numerous notable correlations, these were frequently not as anticipated. As previously stated, the prompts have been constructed based on textual indicators derived from the criteria descriptions. However, only 20\% of the indicators aligned with the criteria for which they were originally constructed. These are the features “comprehensibility” and “uncommon” for the criterion “sociolinguistic appropriateness” and the feature “substitution” for the criterion “coherence and cohesion”. Although five additional features can be identified with a statistically significant correlation of $|PCC|\geq.60$ at a .05 level
, the overall accuracy with regard to the intended criterion remains relatively low. However, such a result does not mean that the indicators were incorrectly processed. Instead, they demonstrate that the high-level criteria are complex combinations that can be challenging to determine. Furthermore, there can be indicators that the teachers implicitly employed but were not part of the criterion. Some indicators could also be specified as a sub-criterion, but given less attention or may be overlooked during the evaluation process. Subsequently, the findings can also help teachers to reflect on the feedback for the criteria considered to reveal which indicators were not given enough consideration or have additionally been evaluated, although they have not been part of the criteria.
Several limitations should be considered regarding the data, the analysis, and the framing factor. Namely, the results may be precisely representative of the data evaluated, and the matching process is subject to rater bias. Data are collected from a relatively small cohort for a specific task, presenting a challenge associated with a limited sample size. Under these circumstances, it is difficult to determine whether the analysis results will generalize to other cohorts and tasks. However, this alone does not invalidate the results. On the contrary, the experimental results provide evidence for the approach's effectiveness. While certain features may theoretically align with the ratings, the presence of data imbalances — such as a majority of students meeting the specified criteria — can result in low correlations. Subsequently, it is essential to identify the feature set with a particular relation to known ratings. Even if they have not been as anticipated, they can serve as a valuable resource for predicting criteria ratings and elucidating the processes through which they might be determined.
Including indirect alignment in the data analysis methods reduces potential adverse outcomes in automated feedback generation. Alignment allows for the validation of generated indicators against teacher-assigned ratings. Since teachers are blind to LLM-generated indicators during the evaluation, the potential for bias and subjectivity in validation results are reduced. The use of indirect alignment has the additional practical benefit of scale since already existing text-rating pairs can be utilized to validate the efficacy of prompts. Therefore, indirect alignment enables the reduction of bias in alignment and provides the opportunity to improve the rigor of the utilized prompts.
However, it must be said that we considered correlations only in this study. There can be features for which no linear relation exists. Exemplarily, a list of known expressions is extracted by the LLM, those count is processed. There could be an optimal number (e.g., three expressions), but including fewer or more items is rated worse. Then, no linear relationship can be identified. Subsequently, non-linear regressions could be included to identify features with such effects. However, the presented study focused on correlations only. Further research can include other forms of non-linear or even mutual regressions.

The methodology of this paper was based on trust and transparency, with ethical data collection and examination. Students were well-informed and felt confident participating. Nevertheless, only a few students submitted texts. As for learners' motivation, collected submissions were the product of a non-compulsory task. Although this language course was an optional choice of students in their curriculum, the language teacher insisted on both parties' learning goals and commitment in the learning agreement: the teacher provided feedback on each written text.

Writing short narratives and repeating structures regularly helps reinforce vocabulary and grammar, which is crucial for other production skills like speaking accurately and fluently. Providing instant and qualified feedback and assisting learners to correct mistakes may improve their productivity, i.e., writing and speaking language skills more effectively as their implicit knowledge is increased \cite[p. 25-28]{Esmaeeli2020EffectWFonoralskills}.
Practices in giving feedback seem to evolve into more effective learning coaching here in our case due to three main changes in the teaching scenario: they affect the rhythm of the writing activities, the order in which the tasks and feedback were performed, and the nature of provided feedback. After receptive activities (listening and reading), learners wrote a text and reviewed the text in pairs after receiving initial explicit corrective feedback. Then, they resubmitted it and obtained implicit formative feedback, i.e., pointing out erroneous passages without correcting them, including advice on further learning steps. Then, they were well prepared for the last productive task: being able to speak in small groups and alone in front of the entire group about one of the topics addressed in the written productions. Some students who participated in this online writing activity reported that it helped them gain self-confidence by writing before speaking, reducing anxiety about lack of vocabulary or structures. Receiving corrective, peer-to-peer, and formative feedback regularly contributed to developing a better feeling for the foreign language as well as motivation to improve their language skills. As to AI-generated feedback in upcoming times, learners' self-regulation and autonomy could be enhanced if students are provided with advice about additional resources that might be useful for further improvements in writing skills.
The primary contribution of the paper is the development of a method for reliably and transparently producing automated formative feedback. The use of LLMs for the extraction of feedback indicators, allows for the processing of large texts, while limiting the potential for hallucinations. The implementation of random forest models allows students and teachers to validate the criteria based on which the model produces a grade. Finally, the method has the potential to greatly reduce bias in feedback generation since the results of many different graders can be used to train a model. Much of the leanings produced by the experiment revolve around the role of LLMs in feedback generation. LLMs thrive when utilized for text processing tasks. The models are able to reliably retrieve indicators from large corpuses of data, based on precise instructions detailing the task at hand. In the absence of precise instructions, or when asked to directly evaluate arbitrary indicators, the reliability of LLMs declines. 

\section{Future Work}
Despite the promising results, further research is needed to assess if the indicator technique generalizes and is reliable at scale. Additionally, the results of the alignment process are subject to evaluator bias. Similar to how human raters can exhibit biases in the form of grading patterns or prejudices, LLMs can also carry various biases, which could even mirror human biases \cite{stureborg2024large}. In light of this, the alignment process may be problematic if biases overlap and remain unrecognized. To this end, bias in teachers and generated feedback may make fruitful further fields of research. The integration of feedback at scale requires the validation of feedback bias. In this regard, a comparison of bias in individual feedback and generated collective feedback, is a topic of particular interest. 

Furthermore, it is of interest to develop the modalities of automated feedback, LLMs may reliably generate. Subsequent to the research, automatic rating of student submissions along a criteria set was integrated into a teacher-supporting tool as part of a university course. A potential future direction would be the development of methodologies which may allow the generation of explicit, written, formative feedback based on extracted indicators, which outlines how learners may improve their submissions.

In addition, applying and evaluating the presented technique in other disciplines is of great interest. LLMs have previously demonstrated promising results in generating feedback scores for linguistic criteria \cite{mizumoto2023exploring}. However, results were poorer when similar strategies were applied to mathematical tasks \cite{lee2024can,gandolfi2024gpt}. Subsequently, the domain dependence of LLM indicator extraction is of great research interest, with the intention of determining the topics of learning analytics where LLMs are applicable. To conclude, future research directions include the development of LLM-based methodologies for producing indicators in adjacent STEM-based fields and generating explicit feedback, in addition to characterizing human and generated grader bias.

\section{Conclusion}
This study has demonstrated that LLMs show significant promise in rating feedback indicators that align with teacher ones applied to students' submissions in a language learning course. The findings suggest that LLMs are particularly capable of rating relevant indicators. However, some indicators do not align. Subsequently, the systematic approach allows further researchers to identify indicators that are worth asking for and those that return non-aligned metrics, which are not worth mentioning due to the potential incapability of the LLM to rate them properly. Identified indicators can greatly enhance the feedback generation process, allowing teachers to focus more on personalized instruction and student engagement. All in all, LLMs possess the ability to extract relevant information from texts, a capability that traditional NLP approaches often lack. However, the filtering mechanism employed in this study serves as a crucial measure to harness the capabilities of LLMs effectively, mitigating the risk of relying uncritically on their outputs.

\section*{Acknowledgments}
This work was supported by the German Federal Ministry of Education and Research (BMBF), grant number 16DHBKI045.

\bibliographystyle{unsrt}  
\bibliography{references}

\end{document}